\documentclass[runningheads]{llncs}
\usepackage{graphicx}
\usepackage{amsmath}
\usepackage{amsfonts}
\usepackage{multirow}
\usepackage{booktabs}
\usepackage{subfig}
\usepackage{bbm}
\usepackage{floatrow}
\newfloatcommand{capbtabbox}{table}[][\FBwidth]

\usepackage[colorlinks=true]{hyperref}

\usepackage{array}
\newcolumntype{P}[1]{>{\centering\arraybackslash}p{#1}}

\begin{document}

% \title{Self-supervised Interpretable Representation Learning on Longitudinal MRIs}
\title{LSOR: Longitudinally-Consistent Self-Organized Representation Learning}

\titlerunning{LSOR}
% \author{Anonymous}
\author{
Jiahong Ouyang\inst{} \and
Qingyu Zhao\inst{} \and
Ehsan Adeli\inst{} \and
Wei Peng\inst{} \and
\\Greg Zaharchuk\inst{}\thanks{co-founder, equity Subtle Medical} \and
Kilian M. Pohl\inst{}\thanks{corresponding author}}
%index{Ouyang, Jiahong}
%index{Zhao, Qingyu}
%index{Adeli, Ehsan}
%index{Peng, Wei}
%index{Zaharchuk, Greg}
%index{Pohl, Kilian}
\authorrunning{J. Ouyang et al.}
% \authorrunning{Anonymous}
% \institute{Anonymous}
\institute{Stanford University, Stanford CA 94305, USA \\}

\maketitle              % typeset the header of the contribution
\begin{abstract}
Interpretability is a key issue when applying deep learning models to longitudinal brain MRIs. One way to address this issue is by visualizing the high-dimensional latent spaces generated by deep learning via self-organizing maps (SOM). SOM separates the latent space into clusters and then maps the cluster centers to a discrete (typically 2D) grid preserving the high-dimensional relationship between clusters. However, learning SOM in a high-dimensional latent space tends to be unstable, especially in a self-supervision setting. Furthermore, the learned SOM grid does not necessarily capture clinically interesting information, such as brain age. To resolve these issues, we propose the first self-supervised SOM approach that derives a high-dimensional, interpretable representation stratified by brain age solely based on longitudinal brain MRIs (i.e., without demographic or cognitive information). Called \textbf{\underline{L}}ongitudinally-consistent \textbf{\underline{S}}elf-\textbf{\underline{O}}rganized \textbf{\underline{R}}epresentation learning (LSOR), the method is stable during training as it relies on soft clustering (vs. the hard cluster assignments used by existing SOM). Furthermore, our approach generates a latent space stratified according to brain age by aligning trajectories inferred from longitudinal MRIs to the reference vector associated with the corresponding SOM cluster. When applied to longitudinal MRIs of the Alzheimer's Disease Neuroimaging Initiative (ADNI, $N=632$), LSOR generates an interpretable latent space and achieves comparable or higher accuracy than the state-of-the-art representations with respect to the downstream tasks of classification (static vs. progressive mild cognitive impairment) and regression (determining ADAS-Cog score of all subjects). The code is available at \url{https://github.com/ouyangjiahong/longitudinal-som-single-modality}.
%\keywords{Self-supervised Learning \and Contrastive Learning  \and Longitudinal Analysis.}
\end{abstract}
\section{Introduction}

% 1. interpretable DL and interpretable latent space
% 2. one way is SOM structuring the latent space as a manifold grid space
% 3. however there is no way to do that for high-dimensional MRI data
% 4. first to generate som for MRI latent representations self-supervised
% 5. moreover, to generate an interpreatble the manifold stratified by the factor of brain aging/cognitive decline, we propose longitudinal regularization

The interpretability of deep learning models is especially a concern for applications related to human health, such as analyzing longitudinal brain MRIs. To avoid interpretation during post-hoc analysis \cite{li2018deep,rudin2019stop}, some methods strive for an interpretable latent representation~\cite{molnar2020interpretable}. One example is self-organizing maps (SOM) \cite{kohonen1990self}, which cluster the latent space so that the SOM representations (i.e., the `representatives of the clusters) can be arranged in a discrete (typically 2D) grid while preserving high-dimensional relationships between clusters. Embedded in unsupervised deep learning models, SOMs have been used to generate interpretable representations of low-resolution natural images \cite{fortuin2018som,manduchi2019dpsom}. 

Intriguing as it sounds, we found their application to (longitudinal) 3D brain MRIs unstable during training and resulted in uninformative SOMs. These models get stuck in local minima so that only a few SOM representations are updated during backpropagation. The issue has been less severe in prior applications\cite{fortuin2018som,manduchi2019dpsom} as their corresponding latent space is of much lower dimension than the task at hand, which requires a high dimension latent space so that it can accurately encode the fine-grained anatomical details in brain MRIs\cite{zhao2020lssl,ouyang2021self}. To ensure all SOM representations can be updated during backpropagation, we propose a soft weighing scheme that not only updates the closest SOM representation for a given MRI but also updates all other SOM representations based on their distance to the closest SOM representation \cite{fortuin2018som,manduchi2019dpsom}. 
% To address the stability issue, we propose a soft weighing scheme to assign each SOM cluster a penalty weight in regularizing the closeness between the SOM representation and a latent representation, so that all SOM representations are updated during backpropagation\cite{fortuin2018som,manduchi2019dpsom}.
Moreover, our model relies on a stop-gradient operator \cite{van2017neural}, which sets the gradient of the latent representation to zero so that it only focuses on updating the SOM representations. It is especially crucial at the beginning of the training when the (randomly initialized) SOM representations are not good representatives of their clusters. Finally, the latent representations of the MRIs are updated via a commitment loss, which encourages the latent representation of an MRI sample to be close to its nearest SOM representation. In practice, these three components ensure stability during the self-supervised training of the SOM on high-dimensional latent spaces.

% To regularize the closeness of the image representation with its best-matching SOM embedding while avoiding being pulled to other SOM embeddingsMoreover, a stop gradient operator is included in the SOM regularization to avoid the image representation being pulled to other SOM embeddings The other two key factors for increasing stability are the stop gradient operator iand an extra commitment loss. Together they regularize the closeness of the representation with its best-matching SOM embedding while avoiding being pulled to other SOM embeddings.

To generate SOMs informative to neuroscientists, we extend SOMs to the longitudinal setting such that the latent space and corresponding SOM grid encode brain aging. Inspired by \cite{ouyang2021self}, we encode pairs of MRIs from the same longitudinal sequence (i.e., same subject) as a trajectory and encourage the latent space to be a smooth trajectory (vector) field. We enforce smoothness by computing for each SOM cluster a reference trajectory, which represents the average aging of that cluster with respect to the training set. The reference trajectories are updated by the exponential moving average (EMA) such that, in each iteration, it aggregates the average trajectory of a cluster with respect to the corresponding training batch (i.e., batch-wise average trajectory). In doing so, the model ensures longitudinal consistency as the (subject-specific) trajectories of a cluster are maximally aligned with the reference trajectory of that cluster.
%To achieve this, we associate each SOM cluster with a trainable reference trajectory representing the average aging direction of that cluster. We then design a longitudinal consistency regularization by enforcing the trajectory (corresponding to a pair of MRIs from the same subject) to be maximally aligned with the reference trajectory of its best-matching SOM cluster. In doing so, a smooth trajectory (vector) field is formed on the latent space to capture the progression of brain aging, and the SOM grid is regularized to be stratified by brain age. During inference, we generate a SOM similarity map for an MRI to estimate the relative brain age by computing the distance of its representation to the SOM embeddings. 

Named \textbf{\underline{L}}ongitudinally-consistent \textbf{\underline{S}}elf-\textbf{\underline{O}}rganized \textbf{\underline{R}}epresentation learning \\(LSOR), %the approach generates the interpretable latent representation of an MRI by computing a 2D similarity grid that stores the similarity scores between the latent representation of that MRI and all SOM representations. As the SOM representations are organized in the 2D grid according to brain age (e.g., brain age is increasing from left to right), each index in the grid represents a brain age and the corresponding entry in the similarity grid (normalized between 0 and 1) encodes the probability of the MRI being of that brain age. Note, the interpretable representation results from self-supervised learning solely relying on longitudinal MRIs, i.e.,  without using any tabular data such as age, cognitive measure, or diagnosis. 
we evaluate our method on a longitudinal T1-weighted MRI dataset of 632 subjects from ADNI to encode the brain aging of Normal Controls (NC) and patients diagnosed with static Mild Cognitive Impairment (sMCI), progressive Mild Cognitive Impairment (pMCI), and Alzheimer's Disease (AD). LSOR clusters the latent representations of all MRIs into 32 SOM representations. The resulting 4-by-8 SOM grid is organized by both chronological age and cognitive measures that are indicators of brain age. Note, such an organization solely relies on longitudinal MRIs, i.e.,  without using any tabular data such as age, cognitive measure, or diagnosis. 
To visualize aging effects on the grid, we compute (post-hoc) a 2D similarity grid for each MRI that stores the similarity scores between the latent representation of that MRI and all SOM representations. As the SOM grid is an encoding of brain aging, the similarity grid indicates the likelihood of placing the MRI within the "spectrum" of aging. Given all MRIs of a longitudinal scan, the change across the corresponding similarity grids over time represents the brain aging process of that individual. Furthermore, we infer brain aging on a group-level by first computing the average similarity grid for an age group and then visualizing the difference of those average similarity grids across age groups. %Visualizing the individual and group aging effects highlights the interpretability of the proposed method regarding brain aging. %Moreover, the smooth trajectory field reveals interpretable aging directions in the latent space. 
With respect to the downstream tasks of classification (sMCI vs. pMCI)  and regression (i.e., estimating the Alzheimer’s Disease Assessment Scale–Cognitive Subscale (ADAS-Cog) on all subjects), our latent representations of the MRIs is associated with comparable or higher accuracy scores than representations learned by other state-of-the-art self-supervised methods.

\section{Method}

\begin{figure}[t]
\centering
\includegraphics[width=0.75\textwidth]{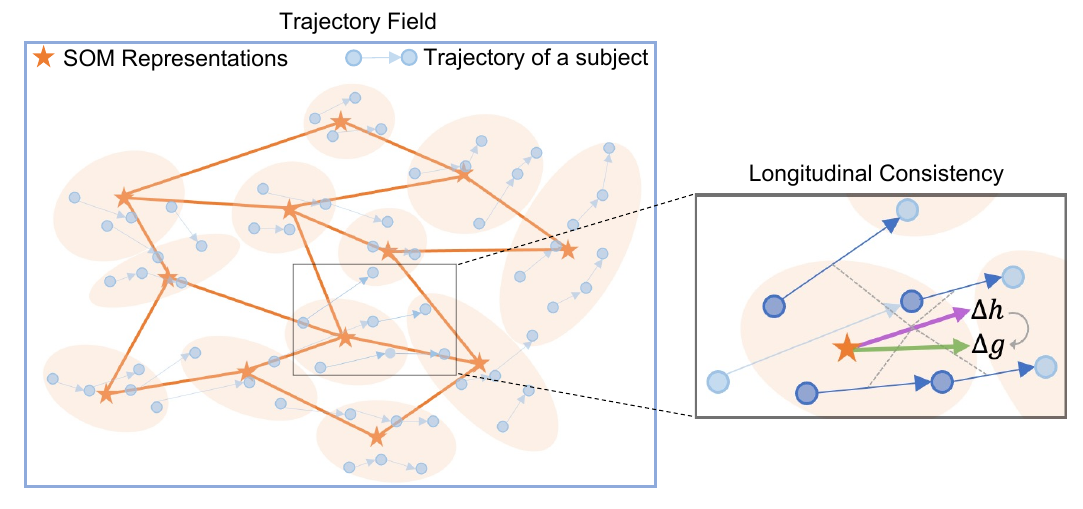}
% \vspace{-10pt}
\caption{Overview of the latent space derived from LSOR. All trajectories ($\Delta z$) form a trajectory field (blue box) modeling brain aging. SOM representations in $\mathcal{G}$ (orange star) are organized as a 2D grid (orange grid). As shown in the black box, reference trajectories $\Delta \mathcal{G}$ (collection of all $\Delta g$, green arrow) are iteratively updated by EMA using the aggregated trajectory $\Delta h$ (purple arrow) across all trajectories of the corresponding SOM cluster within a training batch.
} 
\label{fig:overview}
\end{figure}

As shown in Fig. \ref{fig:overview}, the longitudinal 3D MRIs of a subject are encoded as a series of trajectories (blue vectors) in the latent space. Following ~\cite{ouyang2021self,zhao2020lssl}, we consider a pair of longitudinal MRIs (that corresponds to a blue vector) as a training sample. Specifically, let $\mathcal{S}$ denote the set of image pairs of the training cohort, where the MRIs $x^u$ and $x^v$ of a longitudinal pair $(x^u, x^v)$ are from the same subject and $x^v$ was acquired $\Delta t$ years after $x^u$. For simplicity, $\times$ refers to $u$ or $v$ when a function is separately applied to both time points. The MRIs are then mapped to the latent space by an encoder $F$, i.e., $z^\times:=F(x^\times)$. On the latent space, the trajectory of the pair is denoted as $\Delta z := (z^v - z^u) / \Delta t$, which represents morphological changes. Finally, decoder $H$ reconstructs the input MRI $x^\times$ from the latent representation $z^\times$, i.e., $\tilde{x}^\times:=H(z^\times)$. Next, we describe LSOR, which generates interpretable SOM representations, and the post-hoc analysis for deriving similarity grids.
% cluster the latent space so that the cluster centers can be arranged in a discrete (typically 2D) grid while preserving high-dimensional relationships between the clusters.

\subsection{LSOR} 
% A gradient-based self-organizing map is derived with a loss function, whose gradient is similar to the update rule of the original non-gradient-based SOM. 
Following \cite{fortuin2018som,manduchi2019dpsom}, SOM representations are organized in a $N_r$ by $N_c$ grid (denoted as SOM grid) $\mathcal{G}=\{g_{i,j}\}_{i=1,j=1}^{N_r,N_c}$, where $g_{i,j}$ denotes the SOM representation on the $i$-th row and $j$-th column. This easy-to-visualize grid preserves the high-dimensional relationships between the clusters as shown in by the orange lines in Fig. \ref{fig:overview}. Given the latent representation $z^\times$, its closest SOM representation is denoted as $g_{\epsilon^\times}$, where $\epsilon^\times := argmin_{(i,j)} \parallel z^\times - g_{i,j} \parallel_2$ is its 2D grid index in $\mathcal{G}$ and $\parallel \cdot \parallel_2$ is the Euclidean norm. This SOM representation is also used to reconstruct the input MRI by the decoder, i.e., $\tilde{x}^\times_g=H(g_{\epsilon^\times})$.  To do so, the reconstruction loss encourages both the latent representation $z^\times$ and its closet SOM representation $g_{\epsilon^\times}$ to be descriptive of the input MRI $x^\times$, i.e., 
\begin{equation}
\label{eqn:recon}
L_{recon} := \mathbb{E}_{(x^u, x^v) \sim \mathcal{S}} \left( \sum_{\times \in \{x,v\} }\parallel x^\times - \tilde{x}^\times \parallel_2^2 + \parallel x^\times - \tilde{x}^\times_g \parallel_2^2 \right),
\end{equation}
where $\mathbb{E}$ defines the expected value. The remainder describes the three novel components of our SOM representation.  

\textbf{Explicitly regularizing closeness.} Though $L_{recon}$ implicitly encourages close proximity between $z^\times$ and $g_{\epsilon^\times}$, it does not inherently optimize $g_{\epsilon^\times}$ as $z^\times$ is not differentiable with respect to $g_{\epsilon^\times}$. Therefore, we introduce an additional `commitment' loss explicitly promoting closeness between them:
\begin{equation*}
\label{eqn:commit}
L_{commit} := \mathbb{E}_{(x^u, x^v) \sim \mathcal{S}} \left(\parallel z^u - g_{\epsilon^u} \parallel_2^2 + \parallel z^v - g_{\epsilon^v} \parallel_2^2 \right).
\end{equation*}
%The training requires the presence of $L_{commit}$ because the encoder 

\textbf{Soft Weighting Scheme.} In addition to update $z^\times$'s closest SOM representation $g_{\epsilon^\times}$, we also update all SOM representations $g_{i,j}$ by introducing a soft weighting scheme as proposed in \cite{mulyadi2022xadlime}. Specifically, we design a weight $w^\times_{i,j}$ to regularize how much $g_{i,j}$ should be updated with respect to $z^\times$ based on its proximity to the grid location $\epsilon^\times$ of $g_{\epsilon^\times}$, i.e., 
\begin{equation}
w_{i,j}^\times := \delta \left( e^{-\frac{\parallel \epsilon^\times - (i,j)\parallel_1^2}{2\tau}}\right), 
\label{eqn:weigh}
\end{equation}
where $\delta(w):=\frac{w}{\sum_{i,j} w_{i,j}}$  ensures that the scale of weights is constant during training and $\tau >0$ is a scaling hyperparameter. Now, we design the following loss $L_{som}$ so that SOM representations close to $\epsilon^\times$ on the grid are also close to $z^\times$ in the latent space (measured by the Euclidean distance $\parallel z^\times - g_{i,j} \parallel_2$):
\begin{equation}
L_{som} := \mathbb{E}_{(x^u, x^v) \sim \mathcal{S}} \left(\sum_{g_{i, j} \sim \mathcal{G}}\left( w^u_{i,j} \cdot \parallel z^u - g_{i,j} \parallel_2^2 + w^v_{i,j} \cdot \parallel z^v - g_{i,j} \parallel_2^2 \right)\right).
\label{eqn:som}
\end{equation}

% \begin{equation}
% L_{som} := \mathbb{E}_{(x^u, x^v) \sim \mathcal{S}} \left(\sum_{g_{i, j} \sim \mathcal{G}}\left( w^u_{i,j} \cdot \parallel sg[z^u] - g_{i,j} \parallel_2^2 + w^v_{i,j} \cdot \parallel sg[z^v] - g_{i,j} \parallel_2^2 \right)\right).
% \label{eqn:som}
% \end{equation}

To improve robustness, we make two more changes to Eq. \ref{eqn:som}. First, we account for SOM representations transitioning from random initialization to becoming meaningful cluster centers that preserve the high-dimensional relationships within the 2D SOM grid. We do so by decreasing $\tau$ in Eq. \ref{eqn:weigh} with each iteration so that the weights gradually concentrate on SOM representations closer to $g_{\epsilon^\times}$ as training proceeds: $\tau(t) := N_r \cdot N_c \cdot \tau_{max} \left( \frac{\tau_{min}}{\tau_{max}} \right)^{t/T} $
% \begin{equation*}
% \tau(t) := N_r \cdot N_c \cdot \tau_{max} \left( \frac{\tau_{min}}{\tau_{max}} \right)^{t/T} 
% \end{equation*}
with $\tau_{min}$ being the minimum and $\tau_{max}$ the maximum standard deviation in the Gaussian kernel, and $t$ represents the current and $T$ the maximum iteration. 

The second change to Eq. \ref{eqn:som} is to apply the stop-gradient operator $sg[\cdot]$ \cite{van2017neural} to $z^\times$, which sets the gradients of $z^\times$ to 0 during the backward pass. The stop-gradient operator prevents the undesirable scenario where $z^\times$ is pulled towards a naive solution, i.e., different MRI samples are mapped to the same weighted average of all image representations. This risk of deriving the naive solution is especially high in the early stages of the training when the SOM representations are randomly initialized and may not accurately represent the clusters.

\textbf{Longitudinal Consistency Regularization.} We derive a SOM grid related to brain aging by generating an age-stratified latent space. Specifically, the latent space is defined by a smooth trajectory field (Fig. \ref{fig:overview}, blue box) characterizing the morphological changes associated with brain aging. The smoothness is based on the assumption that MRIs with similar appearances (close latent representations on the latent space) should have similar trajectories. It is enforced by modeling the similarity between each subject-specific trajectory $\Delta z$ with a reference trajectory that represents the average trajectory of the cluster. Specifically, $\Delta g_{i,j}$ is the reference trajectory (Fig. \ref{fig:overview}, green arrow) associated with $g_{i,j}$ then the reference trajectories of all clusters $\mathcal{G}_{\Delta}=\{ \Delta g_{i,j} \}_{i=1,j=1}^{N_r,N_c}$ represent the average aging of SOM clusters with respect to the training set. As all subject-specific trajectories are iteratively updated during the training, it is computationally infeasible to keep track of $\mathcal{G}_{\Delta}$ on the whole training set. We instead propose to compute the exponential moving average (EMA) (Fig. \ref{fig:overview}, black box), which iteratively aggregates the average trajectory with respect to a training batch to $\mathcal{G}_{\Delta}$:
\begin{align*}
\label{eqn:ema}
\Delta g_{i,j} & \leftarrow 
 \begin{cases}
 \Delta h_{i,j} & t=0 \\
 \Delta g_{i,j} & t>0 \mbox{ and } |\Omega_{i,j}| = 0 \\
\alpha \cdot \Delta g_{i,j} + (1-\alpha) \cdot \Delta h_{i,j} & t>0 \mbox{ and } |\Omega_{i,j}| > 0\\
 \end{cases}\\
\mbox{with } \Delta h_{i,j} &:= \frac{1}{|\Omega_{i,j}|} \sum_{k=1}^{N_{bs}} \mathbbm{1}[\epsilon^u_k=(i,j)] \cdot \Delta z_k \mbox{ and } |\Omega_{i,j}|:=\sum_{k=1}^{N_{bs}} \mathbbm{1}[\epsilon^u_k=(i,j)].
\end{align*}
$\alpha$ is the EMA keep rate, $k$ denotes the index of the sample pair, $N_{bs}$ symbolizes the batch size,  $\mathbbm{1}[\cdot]$ is the indicator function, and  $|\Omega_{i,j}|$ denotes the number of sample pairs with $\epsilon^u=(i,j)$ within a batch. Then in each iteration, $\Delta h_{i,j}$ (Fig. \ref{fig:overview}, purple arrow) represents the batch-wise average of subject-specific trajectories for sample pairs with $\epsilon^u = (i,j)$. By iteratively updating $\mathcal{G}_{\Delta}$, $\mathcal{G}_{\Delta}$ then approximate the average
 trajectories derived from the entire training set. Lastly, inspired by \cite{ouyang2021self,ouyang2022self}, the longitudinal consistency regularization is formulated as
\begin{equation*}
\label{eqn:dir}
L_{dir} := \mathbb{E}_{(x^u, x^v) \sim \mathcal{S}} \left( 1 - cos(\theta [\Delta z, sg[\Delta g_{\epsilon^u}]])\right),
\end{equation*}
where $\theta [\cdot, \cdot]$ denotes the angle between two vectors. Since $\Delta g$ is optimized by EMA, the stop-gradient operator is again incorporated to only compute the gradient with respect to $\Delta z$ in $L_{dir}$. 

\noindent\textbf{Objective function.} The complete objective function is the weighted combination of the prior losses with weighing parameters $\lambda_{commit}$, $\lambda_{som}$, and $\lambda_{dir}$:
\begin{equation*}
\label{eqn:final}
L := L_{recon} + \lambda_{commit} \cdot L_{commit} + \lambda_{som} \cdot L_{som} + \lambda_{dir} \cdot L_{dir}
\end{equation*}
The objective function encourages a smooth trajectory field of aging on the latent space while maintaining interpretable SOM representations for analyzing brain age in a pure self-supervised fashion.

\subsection{SOM Similarity Grid} 
During inference, a (2D) similarity grid $\rho$ is computed by the closeness between the latent representation $z$ of an MRI sample and the SOM representations:
\begin{equation*}
\label{eqn:sim}
\rho := softmax(-\parallel z - \mathcal{G} \parallel_2^2 / \gamma) \mbox{ with } \gamma := std(\parallel z - \mathcal{G} \parallel_2^2)
\end{equation*}
$std$ denotes the standard deviation of the distance between $z$ to all SOM representations. As the SOM grid is learned to be associated with brain age (e.g., represents aging from left to right), 
the similarity grid essentially encodes a ``likelihood function" of the brain age in $z$. Given all MRIs of a longitudinal scan, the change across the corresponding similarity grids over time represents the brain aging process of that individual. Furthermore, brain aging on the group-level is captured by first computing the average similarity grid for an age group and then visualizing the difference of those average similarity grids across age groups.

\section{Experiments}
\subsection{Experimental Setting}
\noindent\textbf{Dataset.} We evaluated the proposed method on all 632 longitudinal T1-weighted MRIs (at least two visits per subject, 2389 MRIs in total) from ADNI-1 \cite{petersen2010alzheimer}. The data set consists of 185 NC (age: 75.57 $\pm$ 5.06 years), 193 subjects diagnosed with sMCI (age: 75.63 $\pm$ 6.62 years), 135 subjects diagnosed with pMCI (age: 75.91 $\pm$ 5.35 years), and 119 subjects with AD (age: 75.17 $\pm$ 7.57 years). There was no significant age difference between the NC and AD cohorts (p=0.55, two-sample \textit{t}-test) as well as the sMCI and pMCI cohorts (p=0.75). All MRI images were preprocessed by a pipeline including denoising, bias field correction, skull stripping, affine registration to a template, re-scaling to 64 $\times$ 64 $\times$ 64 volume, and transforming image intensities to z-scores.

\noindent\textbf{Implementation Details.} Let C$_k$ denote a Convolution(kernel size of $3\times3\times3$, Conv$_k$)-BatchNorm-LeakyReLU(slope of 0.2)-MaxPool(kernel size of 2) block with $k$ filters, and CD$_k$ an Convolution-BatchNorm-LeakyReLU-Upsample block. The architecture was designed as C$_{16}$-C$_{32}$-C$_{64}$-C$_{16}$-Conv$_{16}$-CD$_{64}$-CD$_{32}$-CD$_{16}$-CD$_{16}$-Conv$_{1}$, which results in a latent space of 1024 dimensions. The training of SOM is difficult in this high-dimensional space with random initialization in practice, thus we first pre-trained the model with only $L_{recon}$ for 10 epochs and initialized the SOM representations by doing k-means of all training samples using this pre-trained model. Then, the network was further trained for 40 epochs with regularization weights set to $\lambda_{recon}=1.0$, $\lambda_{commit}=0.5$, $\lambda_{som}=1.0$, $\lambda_{dir}=0.2$. Adam optimizer with learning rate of $5 \times 10^{-4}$ and weight decay of $10^{-5}$ were used. $\tau_{min}$ and $\tau_{max}$ in $L_{som}$ were set as 0.1 and 1.0 respectively. An EMA keep rate of $\alpha=0.99$ was used to update reference trajectories. A batch size $N_{bs}=64$ and the SOM grid size $N_r=4, N_c=8$ were applied. 

\begin{figure}[!t]
\centering
\includegraphics[width=0.8\textwidth]{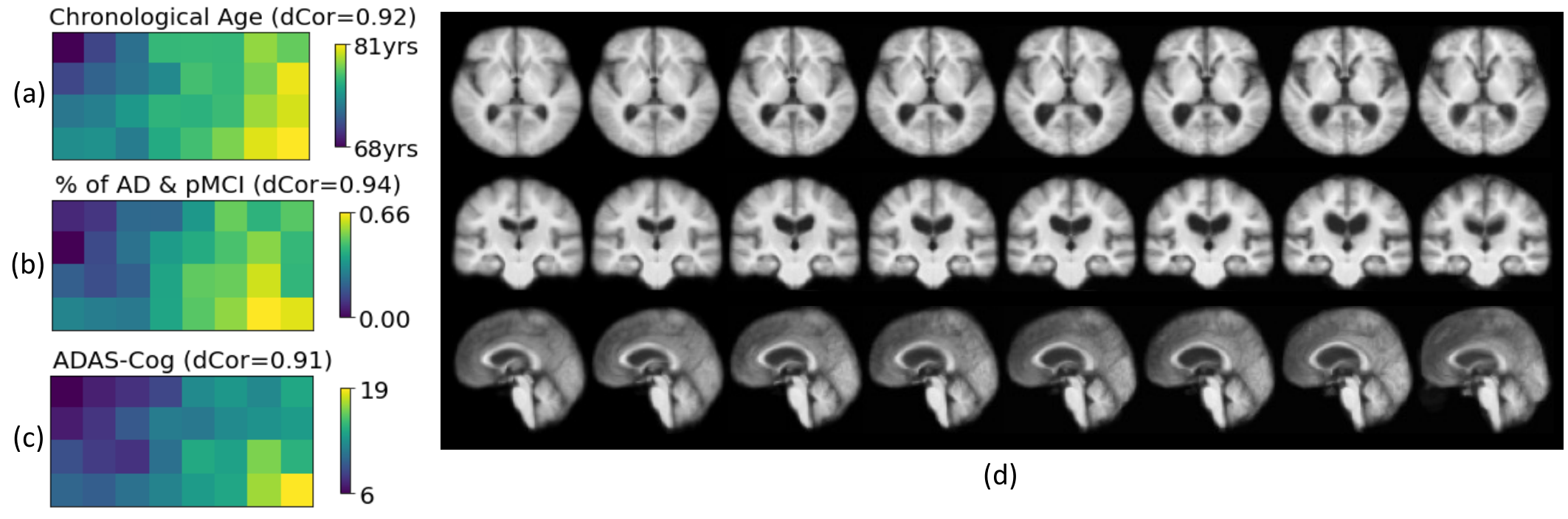}
% \vspace{-20pt}
\caption{The color at each SOM representation encodes the average value of (a) chronological age, (b) \% of AD and pMCI, and (c) ADAS-Cog score across  the training samples of that cluster; (d) Confined to the last row of the grid, the average MRI of 20 latent representations closest to the corresponding SOM representation.} 
\label{fig:som-factor}
\end{figure}

\noindent\textbf{Evaluation.}  We performed five-fold cross-validation (folds split based on subjects) using  10\% of the training subjects for validation. The training data was augmented by flipping brain hemispheres and random rotation and translation. To quantify the interpretability of the SOM grid, we correlated the coordinates of the SOM grid with quantitative measures related to brain age, e.g., chronological age, the percentage of subjects with severe cognitive decline, and Alzheimer’s Disease Assessment Scale–Cognitive Subscale (ADAS-Cog). We illustrated the interpretability with respect to brain aging by visualizing the changes in the SOM similarity maps over time. We further visualized the trajectory vector field along with SOM representations by projecting the 1024-dimensional representations to the first two principal components of SOM representations. Lastly, we quantitatively evaluated the quality of the representations by applying them to the downstream tasks of classifying sMCI vs. pMCI and ADAS-Cog prediction. We measured the classification accuracy via Balanced accuracy (BACC) and Area Under Curve (AUC) and the prediction accuracy via Coefficient of Determination (R2) and root-mean-square error (RMSE). The classifier and predictor were multi-layer perceptrons containing two fully connected layers of dimensions 1024 and 64 with a LeakyReLU activation. We compared the accuracy metrics to models using the same architecture with encoders pre-trained by other representation learning methods, including unsupervised methods (AE, VAE \cite{kingma2013auto}), self-supervised method (SimCLR \cite{chen2020simple}), longitudinal self-supervised method (LSSL \cite{zhao2020lssl}), and longitudinal neighborhood embedding (LNE \cite{ouyang2021self}). All comparing methods used the same experimental setup (e.g., encoder-decoder, learning rate, batch size, epochs, etc), and the method-specific hyperparameters followed \cite{ouyang2021self}.

\begin{figure}[!t]
\centering
\includegraphics[width=0.9\textwidth]{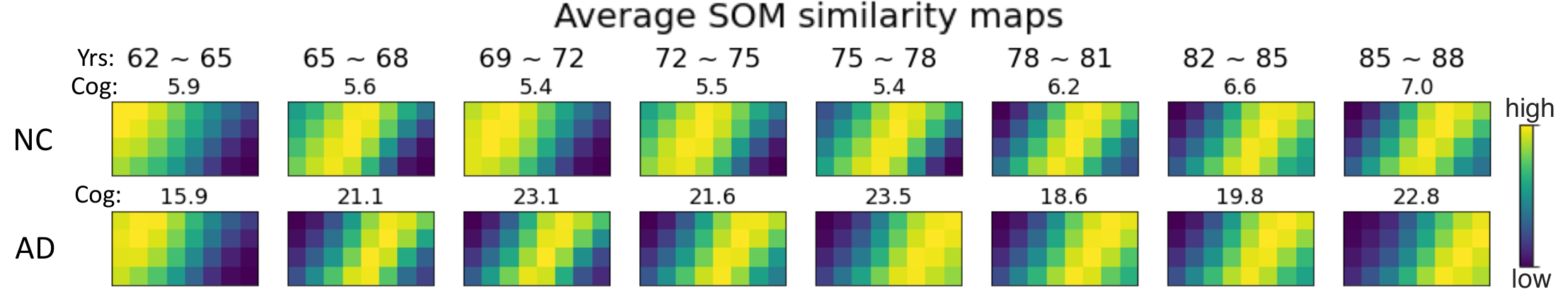}
% \vspace{-20pt}
\caption{The average similarity grid $\rho$ over subjects of a specific age and diagnosis (NC vs AD). Each grid encodes the likelihood of the average brain age of the corresponding sub-cohort. Cog denotes the average ADAS-Cog score.}
\label{fig:som-sim}
\end{figure}

\subsection{Results}
\noindent\textbf{Interpretability of SOM representations.}
Fig. \ref{fig:som-factor} shows the stratification of brain age over the SOM grid $\mathcal{G}$. For each grid entry, we show the average value of chronological age (Fig. \ref{fig:som-factor}(a)), \% of AD \& pMCI (Fig. \ref{fig:som-factor}(b)), and ADAS-Cog score (Fig. \ref{fig:som-factor}(c)) over samples of that cluster. We observed a trend of older brain age (yellow) from the upper left towards the lower right, corresponding to older chronological age and worse cognitive status. The SOM grid index strongly correlated with these three factors (distance correlation of 0.92, 0.94, and 0.91 respectively). Fig. \ref{fig:som-factor}(d) shows the average brain over 20 input images with representations that are closest to each SOM representation of the last row of the grid (see Supplement Fig. S1 for all rows). From left to right the ventricles are enlarging and the brain is atrophying, which is a hallmark for brain aging effects.

\noindent\textbf{Interpretability of similarity grid.} 
Visualizing the average similarity grid $\rho$ of the  NC and AD at each age range in Fig. \ref{fig:som-sim}, we observed that higher similarity (yellow) gradually shifts towards the right with age in both NC and AD (see Supplemental Fig. S2 for sMCI and pMCI cohorts). However, the shift is faster for AD, which aligns with AD literature reporting that AD is linked to accelerated brain aging\cite{toepper2017dissociating}. Furthermore, the subject-level aging effects shown in Supplemental Fig. S3 reveal that the proposed visualization could capture subtle morphological changes caused by brain aging.

\begin{figure}[t!]
\begin{floatrow}
\ffigbox{%
\includegraphics[width=0.4\textwidth]{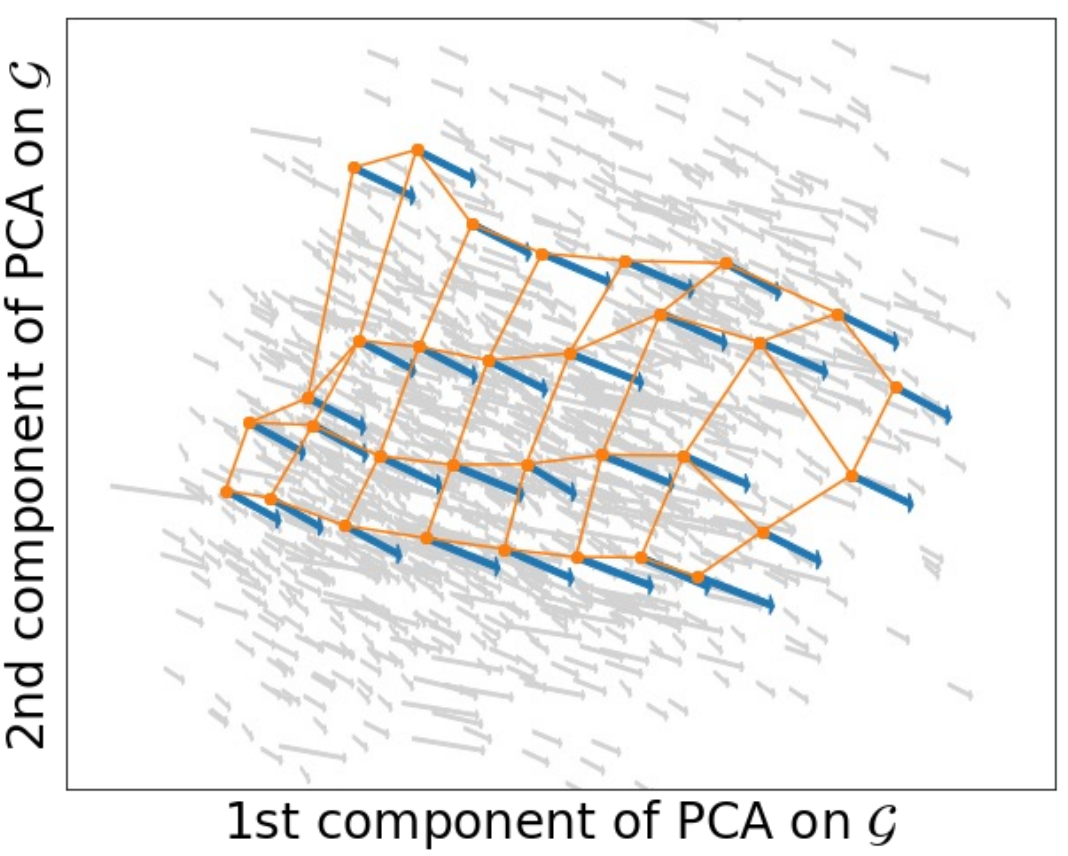}
}{%
  \caption{2D PCA of the LSOR's latent space. Light gray arrows represent $\Delta z$. The orange grid represents the relationships between SOM representations and associated reference trajectory $\Delta \mathcal{G}$ (blue arrow).} 
  \label{fig:som-pca}
}
\capbtabbox{%
    \begin{tabular}
    {c|P{0.9cm}|P{0.9cm}||P{0.9cm}|P{0.9cm}}
        \toprule
        \multirow{2}{*}{Methods} & \multicolumn{2}{c||}{sMCI/pMCI} & \multicolumn{2}{c}{ADAS-Cog}\\
        \cline{2-5}
         & BACC & AUC & R2 & RMSE\\
        \hline
        AE  & 62.6 & 65.4 & 0.26 & 6.98 \\
        VAE \cite{kingma2013auto} & 61.3 & 64.8 & 0.23 & 7.17\\
        SimCLR \cite{chen2020simple} & 63.3 & 66.3 & 0.26 & 6.79\\
        LSSL \cite{zhao2020lssl} & 69.4 & 71.8 & 0.29 & 6.49 \\
        LNE \cite{ouyang2021self} & \textbf{70.6} & 72.1 & 0.30 & 6.46 \\
        % VAE [3] & 61.3 & 64.8 & 0.23 & 7.17\\
        % SimCLR [4] & 63.3 & 66.3 & 0.26 & 6.79\\
        % LSSL [5] & 69.4 & 71.8 & 0.29 & 6.49 \\
        % LNE [6] & \textbf{70.6} & 72.1 & 0.30 & 6.46 \\
        
        % LNE+ \cite{ouyang2022self} & 82.1 & 85.4 & 71.1 & 73.7 & & \\
        \hline
        LSOR & 69.8 & \textbf{72.4} & \textbf{0.32} & \textbf{6.31}\\
    \bottomrule
    \end{tabular}
}{%
  \caption{Supervised downstream tasks using the learned representations $z$ (without fine-tuning the encoder). LSOR achieved comparable or higher accuracy scores than other state-of-the-art self- and un-supervised methods.}%
  \label{tab:res}
}
\end{floatrow}
\end{figure}

\noindent\textbf{Interpretability of trajectory vector field.} 
Fig. \ref{fig:som-pca} plots the PCA projections of the latent space in 2D, which shows a smooth trajectory field (gray arrows) and reference trajectories $\mathcal{G}_{\Delta}$ (blue arrows) representing brain aging. This projection also preserved the 2D grid structure (orange) of the SOM representations suggesting that aging was the most important variation in the latent space.

% \begin{figure}[t]
% \centering
% \includegraphics[width=0.4\textwidth]{fig-som-pca.pdf}
% \vspace{-5pt}
% \caption{Interpretability of the trajectory vector field: latent space of the proposed LSOR projected into 2D PCA space of SOM embeddings $\mathcal{G}$. Arrows represent $\Delta z$ and are color-coded by diagnosis groups. The grid represents the topological structure of SOM embeddings $\mathcal{G}$.} 
% \label{fig:som-pca}
% \end{figure}

% \begin{table}[!t]
% \centering
% \small
% \begin{tabular}
% {c|P{1.5cm}|P{1.5cm}||P{1.5cm}|P{1.5cm}||P{1.5cm}|P{1.5cm}}
%     \toprule
%     \multirow{2}{*}{Methods} & \multicolumn{2}{c||}{NC vs AD} & \multicolumn{2}{c||}{sMCI vs pMCI} & \multicolumn{2}{c}{ADAS-Cog}\\
%     \cline{2-7}
%      & BACC & AUC & BACC & AUC & R2 & RMSE\\
%     \hline
%     AE & 72.2 & 75.4 & 62.6 & 65.4 & 0.26 & 6.98 \\
%     VAE \cite{kingma2013auto} & 66.7 & 70.0 & 61.3 & 64.8 & 0.23 & 7.17\\
%     SimCLR \cite{chen2020simple} & 72.9 & 75.9 & 63.3 & 66.3 & 0.26 & 6.79\\
%     LSSL \cite{zhao2020lssl} & 74.2 & 77.8 & 69.4 & 71.8 & 0.29 & 6.49 \\
%     LNE \cite{ouyang2021self} & \textbf{81.9} & \textbf{83.1} & \textbf{70.6} & 72.1 & 0.30 & 6.46 \\
%     % LNE+ \cite{ouyang2022self} & 82.1 & 85.4 & 71.1 & 73.7 & & \\
%     \hline
%     LSOR & 77.9 & 82.7 & 69.8 & \textbf{72.4} & \textbf{0.32} & \textbf{6.31}\\
% \bottomrule
% \end{tabular}
% \vspace{5pt}
% \caption{Supervised downstream tasks using representation $z$.}
% \label{tab:res}
% \end{table}

\noindent\textbf{Downstream Tasks.}
% The quantitative results on the downstream supervised tasks are shown in Table \ref{tab:res}. 
To evaluate the quality of the learned representations, we froze encoders trained by each method without fine-tuning and utilized their representations for the downstream tasks (Table \ref{tab:res}). On the task of sMCI vs.~pMCI classification (Table \ref{tab:res} (left)), the proposed method achieved a BACC of 69.8 and an AUC of 72.4, a comparable accuracy ($p > 0.05$, DeLong's test) with LSSL \cite{zhao2020lssl} and LNE \cite{ouyang2021self}, two state-of-the-art self-supervised methods on this task. On the ADAS-Cog score regression task, the proposed method obtained the best accuracy with an R2 of 0.32 and an RMSE of 6.31. It is worth mentioning that an accurate prediction of the ADAS-Cog score is very challenging due to its large range (between 0 and 70) and its subjectiveness resulting in large variability across exams \cite{connor2008administration} so that even larger RMSEs have been reported for this task~\cite{ma2021multi}. Furthermore, our representations were learned in an unsupervised manner so that further fine-tuning of the encoder would improve the prediction accuracy.

\section{Conclusion}
In this work, we proposed LSOR, the first SOM-based learning framework for longitudinal MRIs that is self-supervised and interpretable. By incorporating a soft SOM regularization, the training of the SOM was stable in the high-dimensional latent space of MRIs. By regularizing the latent space based on longitudinal consistency as defined by longitudinal MRIs, the latent space formed a smooth trajectory field capturing brain aging as shown by the resulting SOM grid. The interpretability of the representations was confirmed by the correlation between the SOM grid and cognitive measures, and the SOM similarity map. When evaluated on downstream tasks sMCI vs.~pMCI classification and ADAS-Cog prediction, LSOR was comparable to or better than representations learned from other state-of-the-art self- and un-supervised methods. In conclusion, LSOR is able to generate a latent space with high interpretability regarding brain age purely based on MRIs, and valuable representations for downstream tasks. 

\section*{Acknowledgement}
 This work was partly supported by funding from the National Institute of Health (MH113406, DA057567, AA017347, AA010723, AA005965, and AA028840), the DGIST R\&D program of the Ministry of Science and ICT of KOREA (22-KUJoint-02), Stanford's Department of Psychiatry \& Behavioral Sciences Faculty Development \& Leadership Award, and by  Stanford HAI Google Cloud Credit. 
% \noindent\textbf{Acknowledgements}

\bibliographystyle{splncs04}
\bibliography{mybibliography}

\newpage
\section*{Supplementary Files}
\setcounter{table}{0}
\renewcommand{\thetable}{S\arabic{table}}
\setcounter{figure}{0}
\renewcommand{\thefigure}{S\arabic{figure}}

\begin{figure}
\centering
\includegraphics[width=0.7\textwidth]{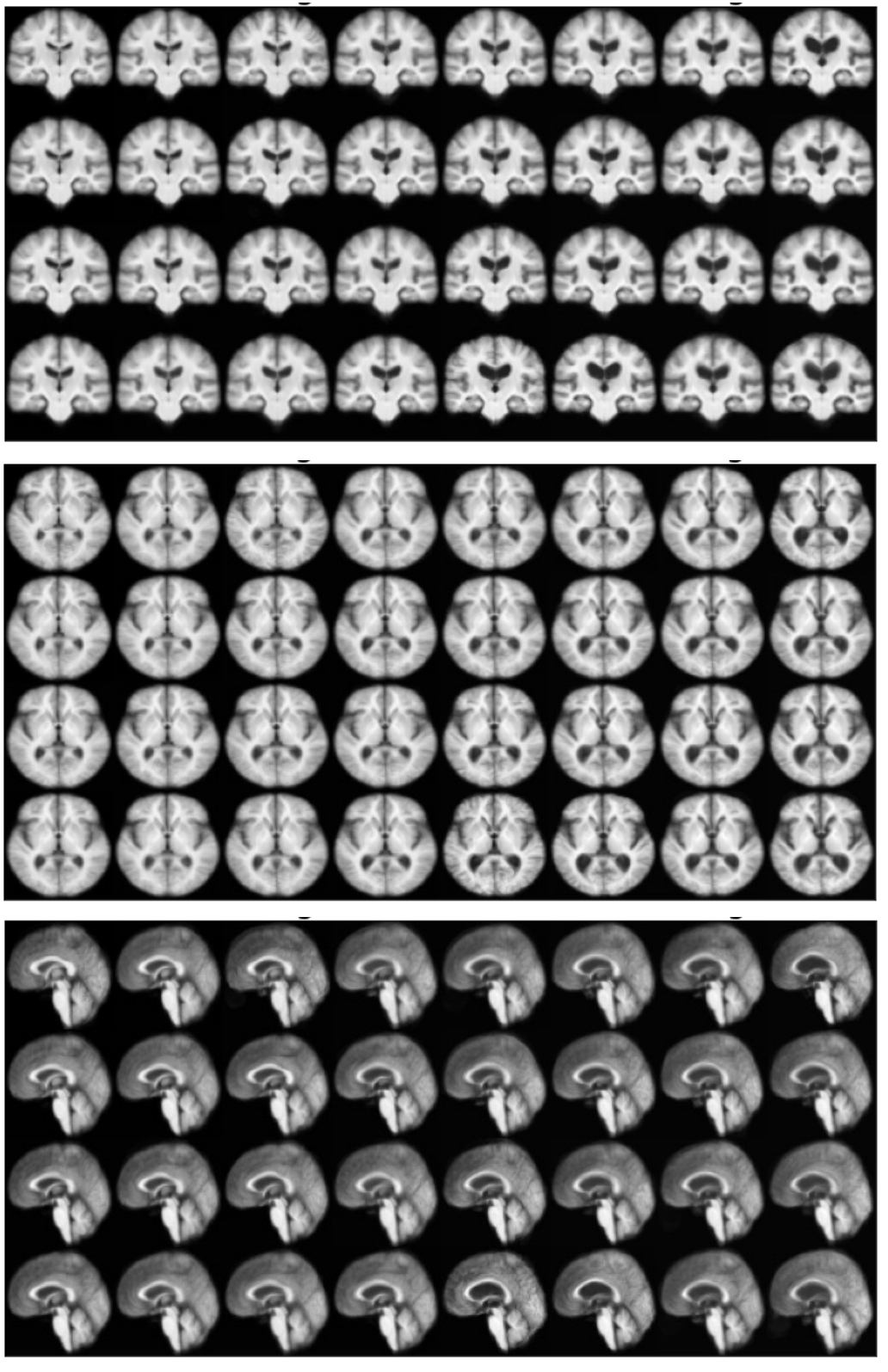}
\vspace{-10pt}
\caption{The average of the 20 nearest samples of each SOM embedding shown in the three views.}
\label{fig:som-img}
\end{figure}

\begin{figure}
\centering
\includegraphics[width=\textwidth]{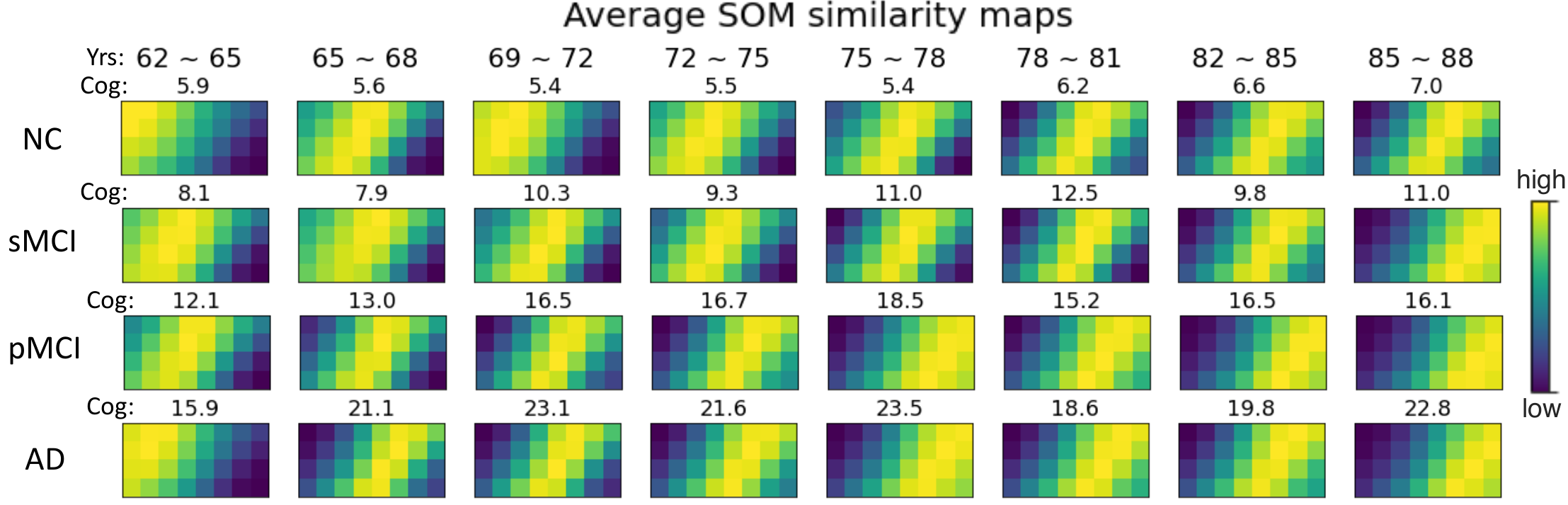}
\vspace{-10pt}
\caption{Interpretability of SOM similarity map $\rho$: the average similarity map for given age groups of NC, sMCI cohort, pMCI, and AD cohort. AD and pMCI cohorts have older brain ages and higher ADAS-Cog scores (Cog) for the same group than NC and AD, illustrated by higher similarity towards the right.}
\label{fig:som-sim-age}
\end{figure}

\begin{figure}[t]
\centering
\includegraphics[width=0.9\textwidth]{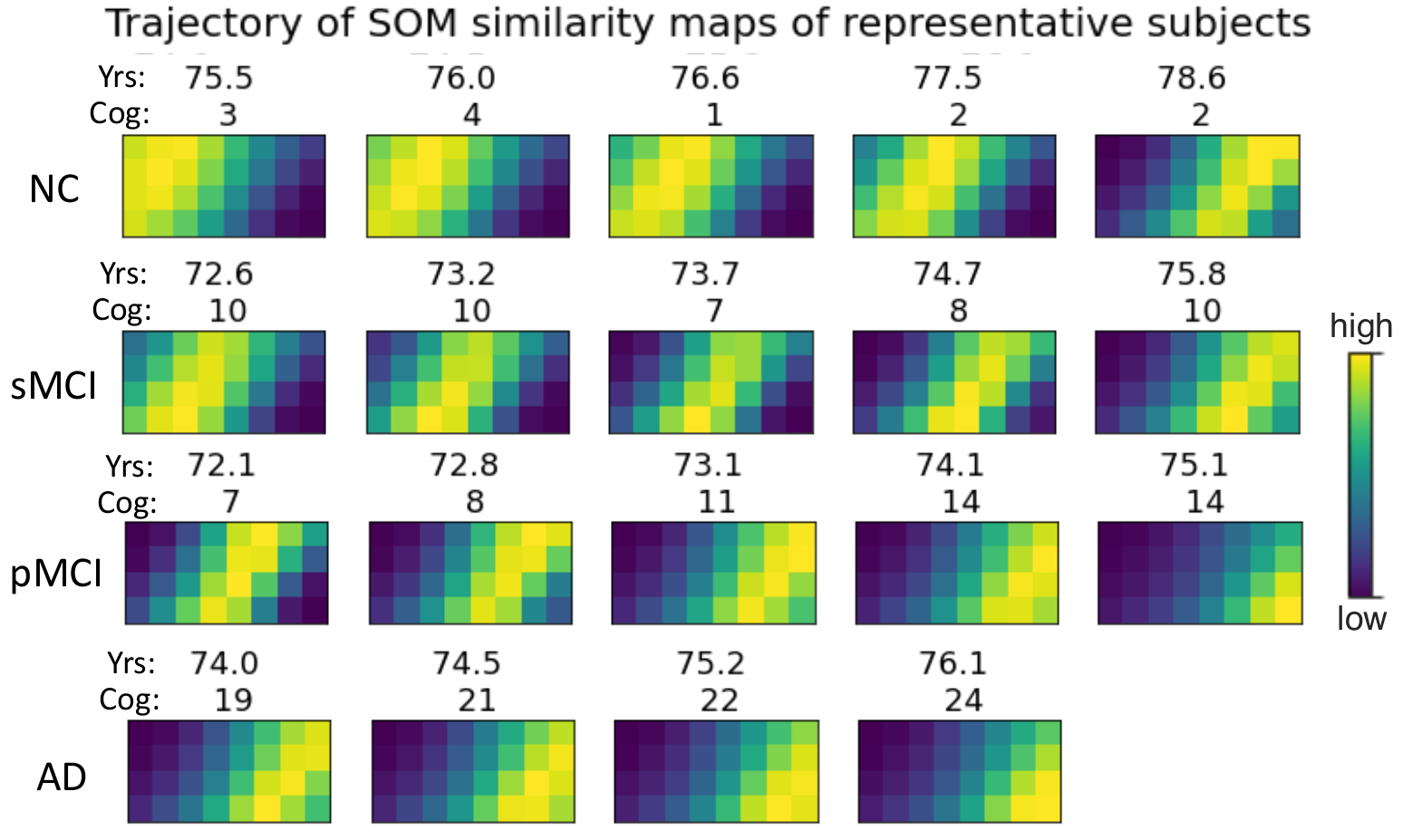}
\vspace{-10pt}
\caption{The trajectory of the SOM similarity map $\rho$ of one representative subject of NC, sMCI, pMCI, and AD cohort. For each subject, the higher similarity gradually moves towards the right as the subject gets older. pMCI and AD subjects tend to have older brain ages and higher ADAS-Cog scores (Cog) than subjects of NC and sMCI at similar chronological ages.}
\label{fig:som-sim-subj}
\end{figure}

\clearpage
\begin{figure}
\centering
\includegraphics[width=1\textwidth]{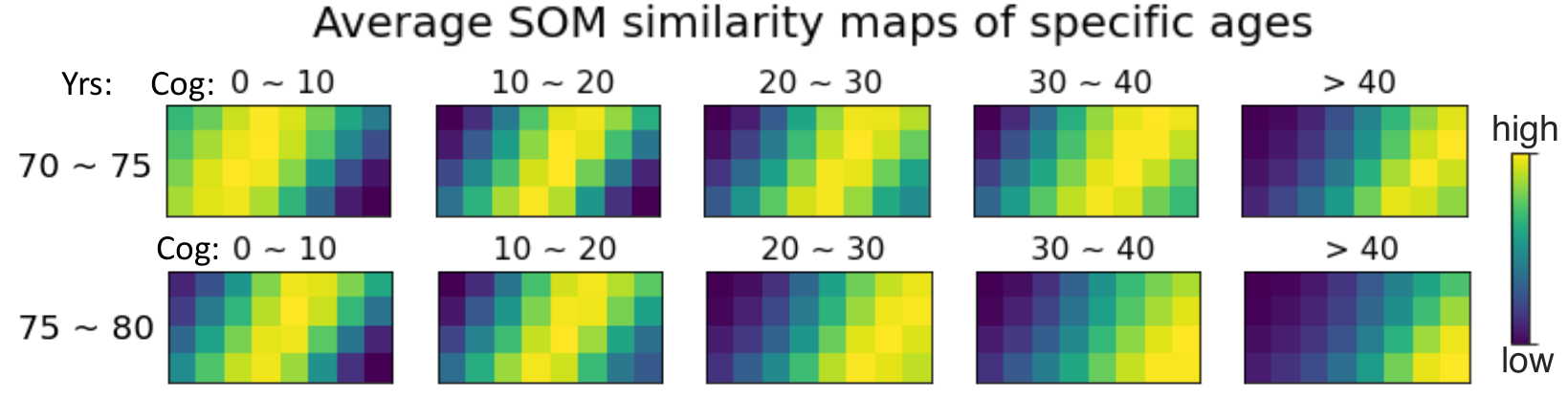}
\vspace{-10pt}
\caption{Interpretability of SOM similarity map $\rho$: the average similarity map for given ADAS-Cog score groups of (a) Age cohort 70-75 years, and (b) Age cohort 70-75 years. For the same age, groups of more severe cognitive impairment (higher ADAS-Cog) have more similarity towards the right.}
\label{fig:som-sim-adas}
\end{figure}

\end{document}